\documentclass[sigconf,nonacm]{style/acmart}
\AtBeginDocument{%
  }

\renewcommand\footnotetextcopyrightpermission[1]{} 

\setcopyright{cc}
\setcctype{by}
\copyrightyear{2026}
\acmYear{2026}

\usepackage{multirow}
\usepackage{wasysym}

\UseMicrotypeSet[expansion]{basicmath}

\newcommand{\fhigh}{\CIRCLE\,}
\newcommand{\fmed}{\LEFTcircle\,}
\newcommand{\flow}{\Circle\,}

\begin{document}

\title{The Trust Paradox: How CS Researchers Engage LLM Leaderboards}

\author{Pouya Sadeghi}
\affiliation{%
  \institution{University of Waterloo}
  \city{Waterloo}
  \country{Canada}
}
\email{pouya.sadeghi@uwaterloo.ca}

\author{Anamaria Crisan}
\affiliation{%
  \institution{University of Waterloo}
  \city{Waterloo}
  \country{Canada}
}
\email{ana.crisan@uwaterloo.ca}

\author{Jimmy Lin}
\affiliation{%
  \institution{University of Waterloo}
  \city{Waterloo}
  \country{Canada}
}
\email{jimmylin@uwaterloo.ca}


\begin{abstract}
Large language model (LLM) leaderboards rank AI models using standardized benchmarks and have become highly visible across computer science, despite known limitations in their reliability and robustness. Yet how they shape researchers' actual practice remains empirically uncharted. We address this gap through semi-structured interviews with eight researchers across four computer science subfields, analyzed using reflexive thematic analysis. We find a near-universal paradox of \emph{pragmatic skepticism}: while participants expressed deep distrust of leaderboard rankings, they continued to use them as rough decision-making aids. Peer networks, not leaderboards, emerged as the primary model selection mechanism, and arena-based (human-voting) leaderboards were consistently preferred over static benchmark leaderboards. Leaderboard influence varied sharply across subfields, revealing that disciplinary culture, not individual attitudes, mediates engagement; for instance, NLP researchers faced state-of-the-art comparison pressure while HCI and Systems/Privacy researchers reported none. Across these differences, however, participants converged on cost transparency as the most demanded missing feature (seven of eight). We translate these findings into concrete design recommendations that align evaluation infrastructure with how researchers actually use it, such as task-specific score breakdowns, cost integration, and voter-demographic disclosure.
\end{abstract}

\keywords{LLM leaderboards, benchmarks, research incentives, socio-technical systems, qualitative
interviews, human–computer interaction.}

\ccsdesc[500]{Human-centered computing~Empirical studies in HCI}
\ccsdesc[500]{Human-centered computing~Human computer interaction (HCI)}
\ccsdesc[300]{Computing methodologies~Artificial intelligence}



\maketitle

\section{Introduction}
\label{sec:intro}

Benchmarks and leaderboards have become central in modern AI research \cite{reducedreusedrecycled2021neurips}. Leaderboards aggregate model scores on standardized tasks and provide continually updated rankings of model score \cite{helm2023tmlr, chatbotarena2024icml}. These leaderboards take two main forms: \emph{static benchmark leaderboards}, such as the Open LLM Leaderboard \cite{openllmleaderboard2023}, MMLU \cite{hendrycks2021mmlu}, which rank models on fixed test sets, and \emph{arena-based leaderboards}, such as Chatbot Arena \cite{chatbotarena2024icml}, which aggregate human preference votes on model outputs. In practice, researchers and practitioners frequently consult LLM leaderboards to evaluate and compare models. For instance, when companies are choosing a model for a task, they often rely on these LLM ranking platforms \cite{droppinghandful2026iclr, votingleaderboardmanipulation2025icml, daynauth2025rankingunraveled, aligningleaderboardwithhumanpreference2025naacl}.

The common assumption is that higher-ranked models represent state-of-the-art (SoTA) capabilities. However, recent work has argued that this assumption may not always hold \cite{measuringwhatmatters2025neurips}. Studies have shown that leaderboards can be sensitive to minor perturbations; for instance, removing a tiny fraction of preference data or tweaking evaluation prompts can change which model is top-ranked \cite{when_benchmarks_are_targets2024acl, droppinghandful2026iclr}. Additionally, it has been shown that swapping ranking algorithms, such as replacing Elo with Glicko, can produce different orderings from the same underlying data \cite{daynauth2025rankingunraveled, amelo2025icml}. Related work also shows that arena rankings can be strategically manipulated through targeted voting, and broader audits report structural concerns in leaderboard operations \cite{voterigging2025icml, leaderboardillusion2025neurips}. Together, these results suggest leaderboard scores are not definitive measures of quality.


Despite the growing body of technical critique, it is still unclear how leaderboard culture shapes researchers' everyday work. If scholars treat leaderboard position as a proxy for progress or suitability, it could narrow research questions or steer model choices; on the other hand, many researchers may distrust or selectively use leaderboard signals. To date, the field lacks empirical evidence about which of these dynamics holds and for whom.

In this project, we adopt a socio-technical lens and treat LLM leaderboards as artifacts embedded in research culture. This perspective aligns with critical analyses that describe AI benchmarks as \textit{deeply political, performative, and generative}; they don’t just measure models, they help shape the field \cite{trustaibenchmarks2025aies}. From an HCI perspective, researchers are end-users of leaderboard platforms whose information needs, trust judgments, and workflow constraints should inform evaluation infrastructure design, yet these needs remain empirically uncharted. Rather than treating leaderboards as purely technical infrastructure, our study centers researchers as people whose practices, trust judgments, and professional norms are shaped by these evaluation platforms. Our qualitative study probes researchers’ lived experiences with leaderboards to see how that shaping actually plays out. Different CS subfields stand in distinct relationships to LLMs, whether as objects of study, tools, or deployment targets, and may engage leaderboards accordingly. By grounding the investigation in interviews across CS subfields, we examine whether leaderboards primarily act as convenient baselines or whether they actively drive research directions and norms.

This study makes three contributions: (1) empirical evidence of a \emph{pragmatic skepticism} paradox, in which researchers distrust leaderboard rankings yet continue using them as rough decision-making aids; (2) a cross-subfield comparison revealing that leaderboard influence is mediated by disciplinary culture rather than individual attitudes; and (3) actionable design recommendations that are grounded in researchers' expressed needs.

The study is organized around two research questions. \textbf{RQ1} asks how researchers across CS subfields use LLM leaderboards when selecting models, designing experiments, and framing research problems. \textbf{RQ2} asks how researchers assess the trustworthiness and limits of leaderboard rankings, and how this assessment varies across leaderboard types (arena-based vs.\ static benchmark).

\section{Related Work}
\label{sec:related_work}

\paragraph{Benchmark Foundations.}
Machine learning communities have long relied on shared benchmarks and public rankings to coordinate progress. Canonical benchmark ecosystems such as ImageNet \cite{deng2009imagenet, russakovsky2015imagenet} and GLUE \cite{wang2018glue} illustrate how common tasks and scoreboards can rapidly focus collective effort. This reflects a broader benchmark culture where state-of-the-art claims shape research agendas \cite{SoTA_benchmark_culture2025springer}.

\paragraph{Benchmark Failure Modes.}
At the same time, a substantial critique of benchmark culture has emerged. \citet{liao2021arewelearningyet} synthesize failure modes (weak baselines, test-set reuse, contamination) and highlight external validity gaps where benchmark gains don't transfer to realistic settings. Interdisciplinary reviews stress transparency, fairness, and explainability in trustworthy evaluation pipelines \cite{trustaibenchmarks2025aies}.

\paragraph{LLM Leaderboard Fragility.}
Work focused specifically on LLM leaderboards reinforces these concerns. \citet{when_benchmarks_are_targets2024acl} show that modest benchmark perturbations, such as answer ordering changes, can shift model rankings substantially. \citet{droppinghandful2026iclr} further show that dropping only a tiny fraction of preference data can flip top-ranked models. Beyond sensitivity, \citet{voterigging2025icml} demonstrate that arena rankings can be strategically manipulated through vote rigging. Governance-focused analyses also report structural issues, including selective disclosure and data-access asymmetries that may distort leaderboard outcomes \cite{leaderboardillusion2025neurips}. Together, rank alone fails to capture broader capability.

\paragraph{Ranking Methods and Statistical Reliability.}
Methodological work has also examined how ranking systems should be designed and interpreted. \citet{aligningleaderboardwithhumanpreference2025naacl} show that automatic LLM ranking pipelines can diverge from human-preference goals depending on component choices. \citet{daynauth2025rankingunraveled} and \citet{amelo2025icml} compare ranking and aggregation strategies, finding that Elo-style implementations can be unstable and that estimation choices materially affect orderings. Related statistical work proposes diagnostic tools and uncertainty quantification for Bradley-Terry models \cite{diagnoseBTframework2022, uncertaintyquantificationBT2023ima}, motivating the reporting of confidence intervals over point ranks.

\paragraph{Socio-Technical Effects of Metrics.}
Beyond robustness, a socio-cognitive literature examines how metrics shape behavior. Goodhart's Law\footnote{In this context, the concern is that once a metric becomes an optimization target, it can lose its value as an indicator of the underlying construct \cite{strathern1997}.} is invoked to explain how leaderboard-driven incentives can warp research priorities \cite{trustaibenchmarks2025aies}. STS analysis adds another layer: \citet{SoTA_benchmark_culture2025springer} describes a present-focused research culture driven by frequent benchmarking, where emphasis can shift toward incremental SoTA gains rather than riskier, long-term innovation.

\paragraph{User Interaction with Algorithmic Rankings.}
HCI research has shown that users often lack awareness of algorithmic curation and develop folk theories to explain ranked outputs. \citet{eslami2015invisiblealgorithms} found that the majority of users were unaware that an algorithm curated their social media feed, and upon discovering this, expressed surprise and re-evaluated their trust. \citet{rader2015userbeliefs} documented a spectrum of user beliefs about how algorithmic curation works, highlighting that ranking opacity shapes trust and behavior. These findings are directly relevant to leaderboard contexts, where researchers encounter opaque rankings and must decide how much to trust them.

\paragraph{Gap and Contribution.}
Prior work treats benchmarks and leaderboards as both technical mechanisms and cultural artifacts. \citet{ethayarajh2020utility} argue that single-metric leaderboard rankings can be poor proxies for the heterogeneous utilities researchers actually care about. Yet there is still limited empirical evidence on how researchers across different CS subfields interpret and use LLM leaderboards in everyday practice. We address that gap through qualitative interviews into researcher attitudes, decisions, and the infrastructural role of leaderboards across subfields.

\section{Methodology}
\label{sec:methods}

This study adopts qualitative methods grounded in an interpretivist epistemology. Our aim is to explore the meanings researchers attribute to leaderboard engagement rather than to generate generalizable causal claims.

\subsection{Ethics and Participant Recruitment}
\label{sec:methods:ethics_recruit}
This study was reviewed and received clearance from the University of Waterloo's human research ethics board. All participants reviewed an information letter and provided informed consent prior to the interview, including separate consent for audio recording and for the use of anonymous quotations in reports and presentations. Participation was voluntary, and participants were free to skip any question or withdraw at any time without consequence. Names, email addresses, and other identifying information were stored separately from the interview data and replaced with participant IDs (P1--P8) during analysis. Audio recordings were anonymized at transcription and stored on a password-protected system.

We recruited researchers from four CS subfields: natural language processing (NLP), Systems/Privacy, human--computer interaction (HCI), and one additional interdisciplinary area (labeled ``Others'' to protect participant anonymity). These subfields were chosen because they represent distinct relationships to LLM leaderboards. NLP researchers are heavy producers and consumers of the benchmarks that feed leaderboards. Systems/Privacy researchers often interact with LLMs as end-users rather than research objects, which brings evaluation perspectives grounded in adversarial thinking. HCI researchers increasingly use LLMs in interactive systems but may prioritize user-centered evaluation criteria that leaderboards typically do not capture. Researchers working in interdisciplinary or domain-specific application areas may apply LLMs to specialized tasks where standard benchmarks have limited relevance.

Inclusion criteria required participants to (a) hold a graduate degree or be enrolled in a graduate program in computer science, (b) have research experience in their respective subfield, and (c) have some familiarity with LLMs, though not necessarily with leaderboards specifically. We recruited through the University of Waterloo CS department and snowball sampling; consequently, some participants were professionally known to the primary researcher prior to the study and others were not. Our final sample consisted of eight researchers (five PhD students, one master's, and two postdoctoral researchers), which provides variation in both disciplinary perspective and career stage. We conducted the eight interviews over two weeks; the final two interviews yielded few new codes, suggesting approaching thematic saturation for the scope of this study.

\subsection{Study Procedure}
\label{sec:methods:procedure}
Each participant went through a three-stage procedure: a pre-interview questionnaire, a semi-structured interview, and a post-interview follow-up.

\paragraph{Pre-interview questionnaire (\textasciitilde5 minutes).} After expressing interest, prospective participants completed a Google Form that combined the consent flow with a short background questionnaire. The form recorded informed consent (including separate consent for audio recording and for anonymous quotation), basic demographic and role information (current position, department/institution, year in program), a brief description of the participant's current research focus, their primary CS subfield, the frequency with which they interact with LLMs or chatbots in their research, prior awareness of named leaderboards (Chatbot Arena, Open LLM Leaderboard, HELM, BigBench, MMLU, SuperGLUE), whether they had ever used a leaderboard or benchmark ranking to inform a research decision, and logistical preferences (video call vs.\ in-person, availability, accessibility needs). The questionnaire allowed us to verify inclusion criteria and to tailor each interview to the participant's familiarity level. The full questionnaire is reproduced in Appendix~\ref{sec:appendix:questionnaire}.

\paragraph{Semi-structured interview (30--45 minutes).} Interviews were conducted over video call or in person, in whichever format the participant had indicated they preferred, and were audio-recorded with explicit consent. To ground the conversation in concrete examples, participants were shown a brief visual typology of leaderboard designs (contrasting arena-based platforms such as Chatbot Arena~\cite{chatbotarena2024icml} with static benchmark platforms such as the Open LLM Leaderboard~\cite{openllmleaderboard2023} and BigBench~\cite{bigbench2023tmlr}) that the research team had prepared from a prior review of leading LLM leaderboard platforms documenting each platform's tasks, metrics, update frequency, and governance model. The interview guide covered familiarity with and attitudes toward LLM leaderboards, concrete use cases such as selecting models or framing research problems, trust and skepticism about leaderboard scores, and reflections on what makes a leaderboard useful or misleading. Representative prompts from each topic area are listed below (the full set appears in Appendix~\ref{sec:appendix:interview_guide}):

\begin{itemize}
    \item \textit{``Tell me about the last time you were choosing a model for a project. What did that process look like?''}
    \item \textit{``Can you walk me through a time when a leaderboard result or benchmark score influenced a decision you made in your research?''}
    \item \textit{``Do you see differences between arena-based rankings like Chatbot Arena and fixed benchmark leaderboards? Do you trust one more than the other?''}
    \item \textit{``In your subfield, how important is it to use the best model according to public rankings?''}
    \item \textit{``If you could redesign the way models are evaluated and compared, what would you change?''}
\end{itemize}

The guide was piloted with two participants and revised based on their feedback before the main data collection; pilot interviews were not included in the analysis reported in this paper, and the eight participants summarised in Table~\ref{tab:participants} are all from the post-pilot main study.

\paragraph{Post-interview follow-up.} After each interview, the primary researcher sent participants a personal thank-you email. As a member-check, each participant was subsequently sent a written summary of their interview and invited to correct factual inaccuracies, clarify intent, or withdraw any quotation. We did not administer a post-interview questionnaire.

\subsection{Data Analysis}
\label{sec:methods:analysis}
All interviews were audio-recorded and transcribed verbatim. We analyzed transcripts using \citet{Braun2006, reflexivethematicanalysis2019qrseh, Braun2021} reflexive thematic analysis (RTA), which we chose because it treats coding as an interpretive, researcher-driven process rather than a reliability exercise, consistent with our interpretivist orientation.

Analysis proceeded through six phases: (1) familiarization with the data, (2) initial code generation, (3) theme searching, (4) theme review, (5) theme definition and naming, and (6) report production. All recruitment, interviewing, coding, theme construction, and write-up were performed by the primary researcher. Consistent with RTA, we did not pursue inter-coder reliability; reflexivity was instead maintained through analytic memo-writing during coding, the participant member-check described in Section~\ref{sec:methods:procedure}, and the positionality considerations stated in Section~\ref{sec:methods:positionality}.

Throughout the analysis, we paid careful attention to variation across subfields and constructed comparative matrices to examine whether themes manifested differently across NLP, Systems/Privacy, HCI, and the Others subfield. The analysis yielded 210 coded segments across 71 unique codes, organized into seven analytical dimensions and consolidated into six themes.\footnote{The seven analytical dimensions are: Model Selection, Leaderboard Engagement, Trust \& Credibility, Vulnerabilities \& Critiques, Missing Metrics, Research Norms, and LLM Usage. The six themes are: Informal Infrastructure, The Social Compass (Peer Networks), Experience Over Evidence, Leaderboards as Academic Currency, Pragmatic Skepticism, and The Evaluation Gap.}

\subsection{Positionality and Reflexivity}
\label{sec:methods:positionality}
Because reflexive thematic analysis treats the analyst's perspective as part of the instrument, we state our positionality explicitly. At the time of the study, the primary researcher (the first author) was a master's student working in natural language processing. He consulted LLM leaderboards regularly both as an object of inquiry and as a working tool in his day-to-day research and engineering practice (for example, when selecting models for coding or deep-search tasks). This dual relationship (analyst-as-user) meant that many of the leaderboard practices participants described were familiar to him from his own experience, which eased rapport during the interviews but also created a risk of anchoring the analysis on his own framings. To mitigate this, codes were generated inductively from the transcripts rather than from a pre-existing framework; analytic memos were kept throughout coding to surface and challenge the researcher's own assumptions; and the member-check described in Section~\ref{sec:methods:procedure} provided participants an opportunity to push back on how their accounts were represented. The authors declare no competing interests, and none of the authors are affiliated with the operators of the leaderboard platforms discussed in this paper.

\subsection{Use of Generative AI}
\label{sec:methods:ai_use}
We used generative AI tools as writing and tooling assistants during the project. Specifically, AI tools were used to (a) edit and polish prose initially drafted by the primary researcher, (b) produce first-pass transcripts of the audio recordings, which the primary researcher verified against the audio during the familiarization phase and again whenever a quotation was selected for inclusion, (c) support literature search, and (d) provide \LaTeX{} scaffolding. Generative AI was \emph{not} used to perform coding, to construct themes, or to interpret the interview data; all analytic decisions were made by the primary researcher. The authors reviewed and edited all AI-generated content and take full responsibility for the final text.

\section{Results}
\label{sec:results}

We conducted semi-structured interviews with eight computer science researchers spanning four subfields: NLP (3), HCI (2), Systems/Privacy (2), and Others (1). Table~\ref{tab:participants} summarizes participant demographics. Thematic analysis yielded 210 coded segments across 71 unique codes, organized into the seven analytical dimensions and six themes described in Section~\ref{sec:methods:analysis}. Rather than the expected spectrum from enthusiastic adoption to skepticism, we found a near-universal stance of pragmatic skepticism, shaped more by social networks and personal experience than by leaderboard rankings.

{\renewcommand{\arraystretch}{1.5}%
\begin{table}[t]
\caption{Participant overview. Each circle represents one participant; fill level indicates self-reported leaderboard familiarity (\fhigh High, \fmed Medium, \flow Low). The sample includes five PhD students, one master's, and two postdocs.}
\Description{A compact table with four rows for NLP, HCI, Systems/Privacy, and Others subfields. Each row shows the number of participants and their leaderboard familiarity as filled, half-filled, or empty circles. NLP has three participants (two high, one medium), HCI has two (one medium, one low), Systems/Privacy has two (both low), and Others has one (medium).}
\label{tab:participants}
\centering
\begin{tabular}{lcl}
\toprule
\textbf{Subfield} & \textbf{n} & \textbf{Leaderboard Familiarity} \\
\midrule
NLP              & 3 & \fhigh \fhigh \fmed \\
HCI              & 2 & \fmed \flow \\
Systems/Privacy  & 2 & \flow \flow \\
Others           & 1 & \fmed \\
\bottomrule
\end{tabular}
\end{table}
}

\begin{figure}[ht]
\centering
\includegraphics[width=0.98\columnwidth]{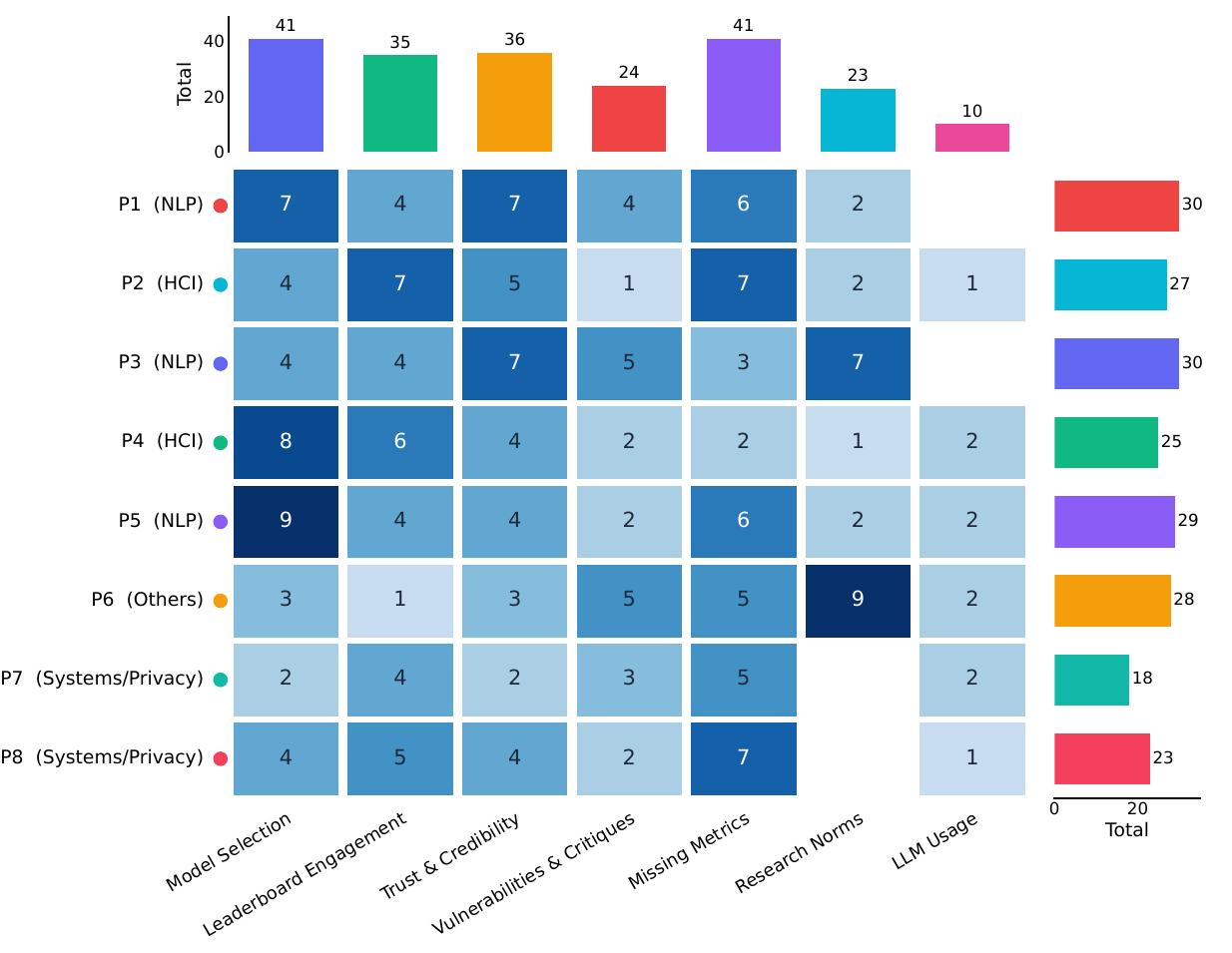}
\caption{Heatmap of raw code citations by participant and analytical dimension. Each cell shows the number of coded segments a participant contributed to a given dimension. Marginal bars summarize row and column totals. NLP participants (P1, P3, P5) show broader engagement across dimensions, while Systems/Privacy participants (P7, P8) concentrate in Model Selection and Missing Metrics.}
\Description{An 8-by-7 heatmap with participants on the vertical axis and analytical dimensions on the horizontal axis. Color intensity indicates the number of raw codes per cell. Marginal bar charts along the top and right edges show column and row totals, respectively. NLP participants have the darkest cells across multiple dimensions; Systems/Privacy participants show concentrated activity in two dimensions.}
\label{fig:heatmap}
\end{figure}

Figure~\ref{fig:heatmap} maps each participant's engagement across the seven analytical dimensions. NLP participants contributed codes broadly, while Systems/Privacy participants concentrated on Model Selection and Missing Metrics, which is consistent with their role as LLM end-users rather than benchmark producers.

\begin{figure*}[t]
\centering
\includegraphics[width=0.99\textwidth]{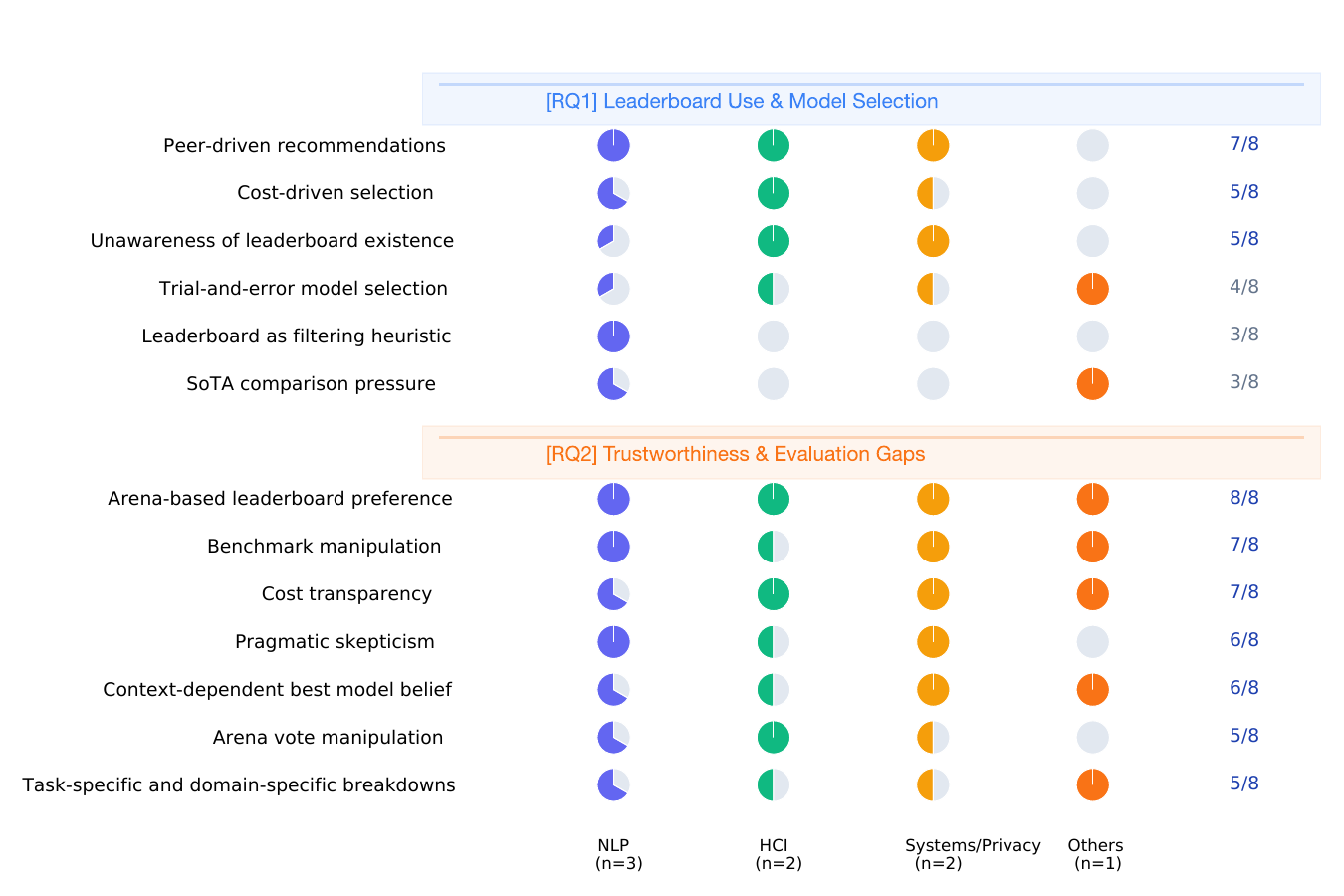}
\caption{Cross-subfield consensus on codes directly relevant to each research question. Pie charts show within-subfield agreement (filled portion = fraction of subfield members who raised the code); the rightmost column shows overall prevalence. Arena-based leaderboard preference achieved full consensus (8/8). Cost transparency and benchmark gaming awareness were raised across all four subfields. SoTA comparison pressure clusters within NLP, while leaderboard unawareness is concentrated in HCI and Systems/Privacy.}
\Description{A dot plot showing codes organized by research question. For each code, pie charts show the proportion of NLP (n=3), HCI (n=2), Systems/Privacy (n=2), and Others (n=1) participants who raised it. Arena-based leaderboard preference is the only code with 8/8 consensus.}
\label{fig:subfield_consensus}
\end{figure*}

Figure~\ref{fig:subfield_consensus} shows code prevalence across subfields for each research question. Arena preference was universal, cost transparency and benchmark manipulations cut all four subfields, while SoTA pressure clustered in NLP, and leaderboard unawareness was concentrated in HCI and Systems/Privacy.

\subsection{RQ1: How Researchers Use Leaderboards in Model Selection}

Four themes capture the strategies, social dynamics, and institutional pressures that shape how researchers select and evaluate LLMs in practice.

\subsubsection{Informal Infrastructure: Navigating Without Formal Guidance}

Perhaps the most surprising baseline finding was that five of eight participants had little to no prior awareness of LLM leaderboards. One HCI researcher said that \textit{``I do not check the model ranking\ldots I've never checked it\ldots not really aware of this concept''}. Both Systems/Privacy researchers, despite deep technical sophistication in their own domains, were similarly unaware of LLM evaluation infrastructure, with one noting they didn't know what the benchmark columns meant upon first seeing them.

Without formal guidance from supervisors or established practices, participants assembled ad-hoc strategies. One HCI researcher consulted vendor API documentation when selecting a model for video analysis. Others relied on brand recognition and defaulted to well-known providers. One Systems researcher picked ChatGPT simply because it was the first model gaining traction among colleagues in the lab. As one HCI participant put it, leaderboards suffer from a discoverability problem: \textit{``there's a huge need\ldots it's just not actually impacting the people it's supposed to impact.''}

\subsubsection{The Social Compass: Peer Networks as Primary Selection Driver}

Seven of eight participants identified peer recommendations as their primary mechanism for selecting LLMs. One NLP researcher quantified this directly: \textit{``90\% of my usage is based on recommendation.''} Channels spanned lab conversations, social media platforms like Twitter/X, and trusted individual experts (\textit{``someone that I know who cares about the quality of work\ldots that is the most important source''}).

Participants expressed an explicit trust hierarchy where peer judgment ranked above all other sources. One NLP researcher stated, \textit{``I trust peers even more than third-party companies, and then the company.''} Vendor claims were systematically discounted: \textit{``each company\ldots they try to promote and advertise it\ldots I can't trust them that much.''} One Systems/Privacy researcher emphasized the need for independent third-party evaluation by arguing that vendor-reported results are inherently unreliable due to incentive misalignment. Several participants also described delayed adoption as a social validation strategy, where they wait for community uptake before trying a new model \cite{rogers2003diffusion}: \textit{``I still take some time and wait till other people start to use that.''}

\subsubsection{Experience Over Evidence: Personal Testing as the Ground Truth}

Seven out of eight participants relied heavily on hands-on trial-and-error testing, consistently prioritizing personal experience over quantitative rankings. Testing took varied forms: running models on personal validation sets (\textit{``we have a validation set, we try everything\ldots whichever has a better result, we would use it''}), quick task-specific probes, or side-by-side comparisons within development tools.

A related pattern was satisficing, where participants accepted models that were ``good enough'' rather than seeking the top-ranked option \cite{simon1955}. One HCI researcher explained, \textit{``maybe that model I'm using is kind of enough for the task\ldots not the best, but enough.''} Cost played a strong role for some: one Systems/Privacy researcher adopted a strictly cost-driven approach, relying on free tiers for most models while maintaining a single paid subscription. Furthermore, regardless of how they initially chose a model, participants exhibited strong path dependency once they found a satisfactory option, routinely reusing it across subsequent projects. Moreover, once a model is good enough and becomes accepted in the literature, it tends to survive; as one NLP researcher noted, even older architectures persist in research pipelines: \textit{``simple Llama 1 or Llama 2 architecture is still being used without switching\ldots doesn't matter.''}

\subsubsection{Leaderboards as Academic Currency: Differential Publication Pressure}

Leaderboard-driven norms varied sharply across subfields. All three NLP participants reported that peer reviewers expect comparisons against state-of-the-art models, echoing broader critiques of SoTA-chasing incentives \cite{lipton2019troubling}. One described it as \textit{``a tradition that you have to compare your methods with highest SOTA,''} while another noted that literature-driven baselines reinforced this pressure: \textit{``I'm heavily biased with the literature because I need to compare.''}

HCI researchers faced no such expectation. One explained, \textit{``HCI papers kind of accept this latency\ldots models used in that paper are no longer the newest.''} Systems/Privacy researchers similarly reported no leaderboard-driven publication pressure; their fields evaluate contributions on theoretical guarantees (privacy bounds, formal proofs) rather than benchmark scores.

Additionally, one researcher highlighted resource inequality: \textit{``I don't think it's fair\ldots I don't have the resources that a researcher in Google would have.''} Reviewers in their field sometimes questioned why they didn't use larger models, despite the practical impossibility of doing so with academic compute budgets. One NLP researcher also described seeking gaps in leaderboard coverage rather than competing on saturated benchmarks: \textit{``novel research comes from small areas they're not doing well.''}

\subsection{RQ2: How Researchers Assess Leaderboard Trustworthiness}

Two themes capture how researchers evaluate leaderboard credibility and what they find missing.

\subsubsection{Pragmatic Skepticism: The Central Paradox of Use Despite Distrust}

\begin{figure}[t]
\centering
\includegraphics[width=0.98\columnwidth]{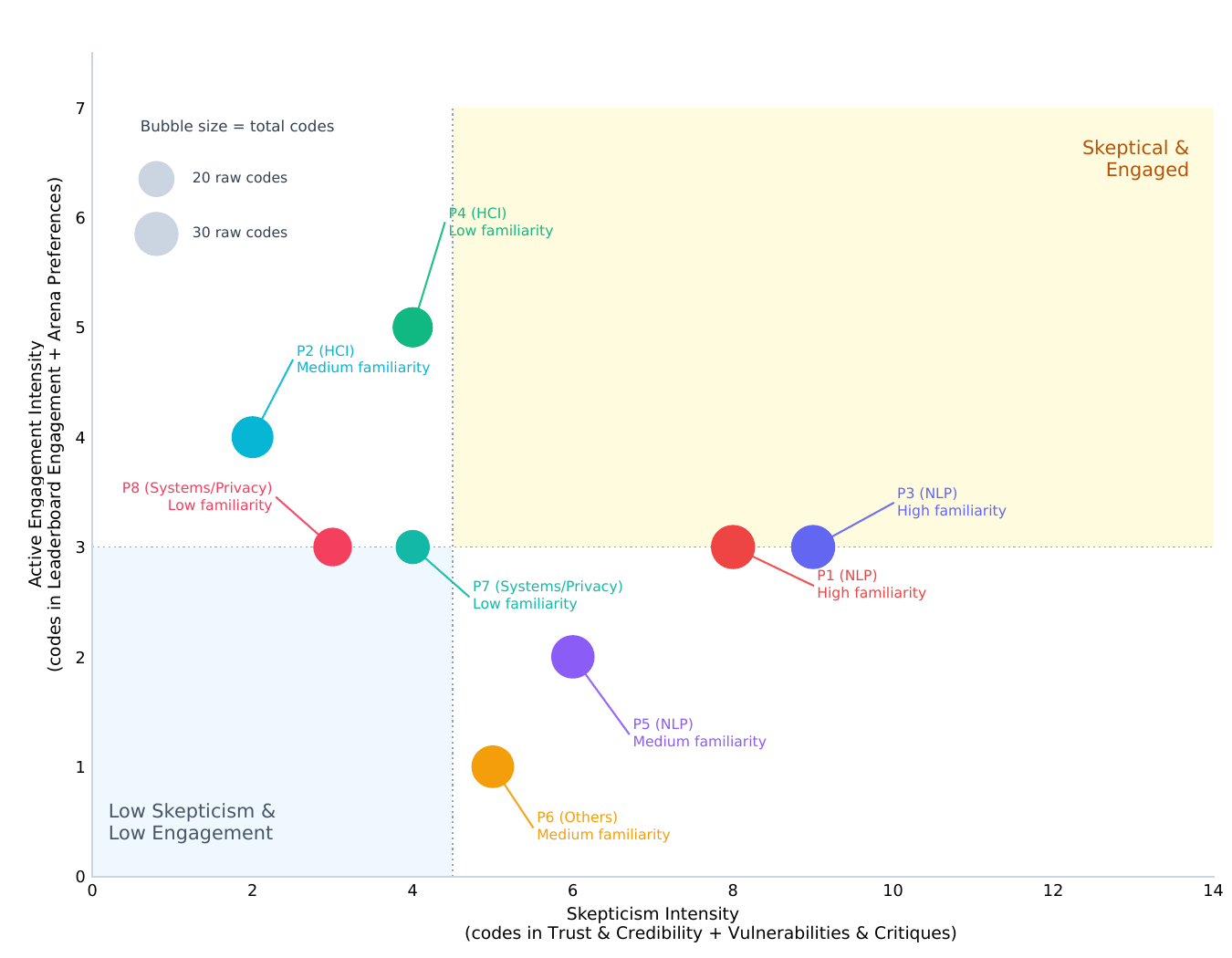}
\caption{The trust paradox: participants simultaneously express skepticism about leaderboard reliability (left axis) and continued engagement with leaderboard rankings (right axis). All eight participants exhibited this paradoxical pattern, treating rankings as ``weak priors'' for narrowing search space rather than as authoritative judgments.}
\Description{A visualization showing the relationship between skepticism indicators and engagement indicators for each participant. All eight participants appear in both the skepticism and engagement regions, illustrating the central paradox that researchers use leaderboards despite distrusting them.}
\label{fig:trust_paradox}
\end{figure}

The central finding was a paradox: all eight participants expressed skepticism about leaderboard trustworthiness yet continued to engage with them as rough decision-making aids (Figure~\ref{fig:trust_paradox}). One NLP researcher described a recurring hype cycle: \textit{``every time a new model is released, everyone is like wow\ldots then disappointed\ldots always a hype.''}

Skepticism was grounded in specific manipulation awareness. Seven of eight participants recognized that static benchmark leaderboards can be gamed: \textit{``fixed leaderboards can be learned\ldots companies can develop their model to outperform on those fixed benchmarks.''} One specifically referenced the Llama~4 arena incident, in which Meta was found to have tested numerous private model variants on Chatbot Arena in the lead-up to the Llama~4 release and submitted a chat-tuned experimental variant that differed substantially from the publicly released model~\cite{leaderboardillusion2025neurips}. Systems/Privacy researchers brought a distinctive flavor of skepticism: one raised concerns about data contamination (models trained on the very test sets used for evaluation), while the other trusted the ranking algorithms' mathematical robustness but distrusted the benchmark data feeding them.

Despite this awareness, participants didn't abandon leaderboards. They treated rankings as weak priors, useful for narrowing the search space but never sufficient for final decisions: \textit{``it is like a prior to somehow limit my search space.''} All eight endorsed the view that there is no universally ``best'' model, and several adopted delayed adoption strategies, waiting weeks or months before trying newly top-ranked models.

\subsubsection{The Evaluation Gap: Misalignment Between Metrics and Researcher Needs}

The most detailed theme concerned the gap between leaderboard metrics and researcher needs.

\paragraph{Arena versus benchmark preference.} All eight participants preferred arena-based (human-voting) leaderboards over static benchmark leaderboards, the only code to achieve complete consensus across all four subfields. Reasons included greater interpretability (\textit{``the arena one is more reliable because it's actually the user prompt''}), adaptiveness to evolving preferences, and alignment with actual use experience. For participants outside NLP, benchmark metrics were opaque: \textit{``I do not understand what it means if MMLU is 68\% or 72\%''} \cite{hendrycks2021mmlu}. Two participants proposed hybrid systems combining human voting with benchmark scores.

\paragraph{Missing metrics.} Cost transparency was the most demanded missing feature, raised by seven of eight participants, echoing broader calls to foreground financial and efficiency costs in model evaluation \cite{strubell2019energy, schwartz2020green}. One NLP researcher called it \textit{``the most important one.''} Beyond cost, participants requested confidence intervals to distinguish genuinely different rankings, failure rate and hallucination metrics, and environmental efficiency measures. Systems/Privacy researchers introduced additional demands not raised by other subfields: adversarial robustness benchmarks, latency metrics, model provenance information, and deployment details.

\paragraph{Granularity and transparency.} Five of eight participants wanted task-specific breakdowns rather than single aggregate scores. One NLP researcher wanted \textit{``separate scores for math, coding, multilingual''} instead of a single number. Another researcher pointed out that even coarse categories obscure meaningful variation: \textit{``each category can have subcategories\ldots what do they mean by vision?''} Participants also highlighted unmeasured experiential qualities (conciseness, authenticity, output style) that strongly influence model preference but appear on no leaderboard.

On the transparency front, four participants demanded voter demographic disclosure (\textit{``I need to know what those humans are\ldots who's actually giving those ratings''}), with one Systems/Privacy researcher raising specific concerns about regional and cultural bias in voting populations. One participant raised concerns about institutional conflicts of interest in leaderboard governance, and another proposed a qualitative review format alongside numerical rankings.

\section{Discussion}
\label{sec:discussion}

We anticipated a spectrum of attitudes toward leaderboards, ranging from enthusiastic adoption among ML-heavy researchers to skepticism from those in adjacent fields. Instead, skepticism turned out to be nearly universal. The more revealing question was not \emph{who} trusts leaderboards, but \emph{why use persists despite distrust}. This paradox, together with the dominance of social trust and notable gaps in leaderboard coverage, reshapes how we should think about the role of these platforms in research practice.

\subsection{Leaderboards as Socially Embedded Artifacts}

Our finding that seven of eight participants rely primarily on peer recommendations supports a socio-technical reading of these platforms. Leaderboards are designed as objective, centralized evaluation tools, yet in practice they operate within social networks of trust where rankings serve as one input among many, filtered through colleagues, Twitter/X discourse, and personal experience. This aligns with prior critiques that benchmark culture shapes research agendas through social mechanisms rather than purely technical ones \cite{SoTA_benchmark_culture2025springer}. The trust hierarchy we observed (peers over third-party systems over vendors) was consistent across all four subfields, and it held even for Systems/Privacy researchers whose own work involves formal trust models. This suggests the social embedding of model selection is deeply ingrained rather than a product of unfamiliarity, and that leaderboard designers can't improve adoption through better metrics alone; they must also address the social credibility of the platform itself.

\subsection{The Pragmatic Skepticism Paradox}

The central paradox (use despite distrust) echoes Goodhart's Law \cite{strathern1997}: once leaderboard scores become targets, they cease to be reliable measures. Our participants articulated this clearly by recognizing that static benchmarks motivate gaming and that arena-based platforms are vulnerable to vote manipulation. Despite these limitations, they continued to use leaderboards as ``weak priors,'' where they treated them as rough filters rather than ground truth. This pragmatic stance implies that the value of leaderboards lies not in their accuracy but in their function as coordination devices and shared reference points that reduce an otherwise overwhelming search space. Systems/Privacy researchers enriched this finding with domain-specific nuance: one distinguished between trusting the mathematical robustness of ranking algorithms and distrusting the benchmark data that feeds them, suggesting that skepticism is not monolithic but structurally differentiated. At the same time, the rapid model release cadence may be eroding the credibility leaderboards depend on, as participants described fatigue with the release-hype-disappointment cycle, which resonates with prior critiques of SoTA-driven incentives \cite{lipton2019troubling}.

\subsection{Design Implications}

Our findings point to concrete design recommendations for leaderboard platforms, which empirically reinforce utility-theoretic critiques that single-score leaderboards often fail to capture what researchers actually need \cite{ethayarajh2020utility}:

\paragraph{Cost transparency} emerged as the single most demanded feature (seven of eight participants). Current leaderboards focus almost exclusively on capability metrics, yet cost-per-query is often the binding constraint for academic researchers \cite{strubell2019energy, schwartz2020green}. Integrating pricing information, or at least efficiency-normalized performance scores, would make leaderboards substantially more useful for practical decision-making.

\paragraph{Arena-based evaluation with statistical rigor.} Arena-based rankings were universally preferred over static benchmark leaderboards, which is consistent with recent work on Bradley-Terry diagnostics and uncertainty quantification \cite{diagnoseBTframework2022, uncertaintyquantificationBT2023ima}. However, participants also recognized arena vulnerabilities and suggested that hybrid systems combining human judgment with standardized benchmarks may offer the best balance. Pairing this with confidence intervals would help researchers distinguish genuinely different models from those within noise margins; this addresses the ranking instability documented in prior work \cite{daynauth2025rankingunraveled, amelo2025icml}.

\paragraph{Task-specific granularity} was requested by five of eight participants. Aggregate scores obscure domain-level variation that researchers care about: a model's coding ability tells a domain-specific researcher little about its performance on specialized scientific tasks. Leaderboards should expose fine-grained, domain-specific breakdowns rather than single composite rankings.

\paragraph{Transparency, provenance, and onboarding.} Participants wanted to know who is voting and what their qualifications are, who built a model and what institutional affiliations exist, and whether conflicts of interest are present. One Systems/Privacy researcher raised specific concerns about regional and cultural bias in voting populations. These are tractable design problems: requiring voter demographics, disclosing funding relationships, and publishing evaluation methodology in machine-readable form would directly address these concerns, consistent with broader calls for structured transparency reporting \cite{mitchell2019modelcards}. A Systems researcher who encountered leaderboards for the first time also proposed onboarding documentation (a user manual explaining methodology and interpretation guidelines) to make these platforms accessible to researchers outside the NLP benchmark ecosystem.

\subsection{Subfield-Specific Implications}

The sharp contrast between NLP's SoTA-comparison norm and its absence in HCI, Systems/Privacy, and the Others subfield reveals that leaderboard influence is mediated by disciplinary culture, not just individual attitudes \cite{lipton2019troubling}. Publication norms that penalize non-top-ranked models create an uneven playing field, particularly for researchers without industry-scale compute \cite{strubell2019energy, schwartz2020green}. One participant, for instance, described being asked by reviewers why a larger model had not been used despite lacking the computing resources to do so; this shows how leaderboard culture can encode resource inequality into the review process itself. Meanwhile, the finding that five of eight participants were unaware of leaderboards' existence points to a fundamental discoverability failure: if these platforms aim to serve the broader research community, they need to reach beyond the NLP audience that currently dominates their user base. Systems/Privacy researchers also brought unique concerns that general-purpose leaderboards don't address; together, these findings suggest that current platforms are designed primarily for and by the NLP community.

\subsection{Ethical Considerations}

Our findings reveal several ethical dimensions of leaderboard culture that warrant attention. First, the widespread awareness of benchmark gaming (seven of eight participants) raises concerns about the integrity of model evaluation: if companies can systematically optimize for leaderboard metrics without corresponding improvements in real-world capability, rankings become misleading signals that distort both research priorities and downstream deployment decisions. This is particularly consequential when practitioners outside NLP (who, as our study shows, may lack the expertise to critically evaluate benchmark methodology) rely on these rankings to select models for applied settings.

Second, the resource inequality highlighted by one participant represents a structural equity concern. When publication norms expect comparisons against the latest top-ranked models, researchers without industry-scale compute budgets are placed at a systematic disadvantage. This dynamic risks concentrating AI research influence among well-resourced institutions and excluding voices from less-resourced universities and the Global South, reinforcing existing power asymmetries in the field \cite{strubell2019energy}.

Third, the lack of transparency in leaderboard governance, such as voter demographics, institutional conflicts of interest, and data-access asymmetries, raises questions about accountability. If leaderboards shape which models are adopted in high-stakes domains such as healthcare or education, the absence of auditable evaluation processes becomes an ethical concern that extends beyond the research community to the broader public.

Finally, regarding the ethics of this study itself: this study was reviewed and received clearance from the University of Waterloo's human research ethics board. All participants provided informed consent, interviews were anonymized during transcription, and no identifying information is reported.

\subsection{Limitations and Future Work}

Our sample of eight participants, while sufficient for identifying themes in small-sample qualitative work \cite{guest2006interviews, Braun2021}, doesn't support claims of generalizability. Each subfield contains only one to three participants, limiting within-subfield variation and precluding strong claims about disciplinary differences. Self-report interview data also captures perceptions and stated practices rather than directly measuring behavioral effects on publication patterns or funding decisions.

Future work should extend this inquiry through larger-scale surveys to quantify the prevalence of the patterns we identified, behavioral studies tracking how leaderboard exposure affects actual model selection, and controlled experiments evaluating the design interventions our participants proposed.

\section{Conclusion}
\label{sec:conclusion}

This study investigated LLM leaderboards as socio-technical artifacts through semi-structured interviews with eight researchers across four CS subfields. Rather than the expected spectrum from adoption to skepticism, we found a pervasive paradox: all participants distrusted leaderboard rankings yet continued using them as rough decision-making aids. Peer networks, not leaderboards, drove model selection, and leaderboard influence varied sharply across subfields.

Our findings yield three primary contributions. First, empirical evidence of a \emph{pragmatic skepticism} paradox, in which researchers distrust leaderboard rankings yet continue to use them as rough decision-making aids that reduce search space and function as shared reference points. Second, a cross-subfield comparison showing that leaderboard influence is mediated by disciplinary culture rather than individual attitudes: state-of-the-art comparison pressure clusters in NLP while HCI, Systems/Privacy, and adjacent subfields report no such expectation. Third, a set of actionable design recommendations, such as cost integration, task-specific granularity, voter transparency, and onboarding documentation, grounded in researchers' expressed needs rather than assumptions about what evaluation should look like. We also identified ethical concerns around resource inequality and benchmark manipulation that leaderboard platforms should address as they evolve.

\begin{acks}
We are grateful to the eight researchers who generously shared their time and reflections during the interviews.
We also thank the instructional team of CS\,889 (Winter 2025) under whose course umbrella the initial phase of the study was conducted.
\end{acks}

\section*{Data Availability}
De-identified interview transcripts are not publicly shareable under the terms of the informed consent that participants provided.
The interview guide, the pre-interview questionnaire, the codebook, and the analytical matrices that support the figures and counts reported in this paper are available from the corresponding author on reasonable request.

\bibliographystyle{ACM-Reference-Format}
\bibliography{main}

@String{Computing = "Computing" }

@String{Computer = "{IEEE} Computer" }

@String{Springer = "Springer-Verlag" }

@inproceedings{votingleaderboardmanipulation2025icml,
  author    = {Huang, Yangsibo and Nasr, Milad and Angelopoulos, Anastasios Nikolas and Carlini, Nicholas and Chiang, Wei-Lin and Choquette-Choo, Christopher A. and Ippolito, Daphne and Jagielski, Matthew and Lee, Katherine and Liu, Ken and Stoica, Ion and Tram\`{e}r, Florian and Zhang, Chiyuan},
  title     = {Exploring and Mitigating Adversarial Manipulation of Voting-Based Leaderboards},
  booktitle = {Proceedings of the 42nd International Conference on Machine Learning},
  series    = {Proceedings of Machine Learning Research},
  volume    = {267},
  pages     = {25654--25671},
  year      = {2025},
  month     = jul,
  publisher = {{PMLR}},
  url       = {https://proceedings.mlr.press/v267/huang25z.html}
}

@inproceedings{when_benchmarks_are_targets2024acl,
  author    = {Alzahrani, Norah A. and Alyahya, Hisham Abdullah and Alnumay, Yazeed and Alrashed, Sultan and Alsubaie, Shaykhah and Almushayqih, Yousef and Mirza, Faisal and Alotaibi, Nouf and Al-Twairesh, Nora and Alowisheq, Areeb and Bari, M. Saiful and Khan, Haidar},
  title     = {When Benchmarks are Targets: Revealing the Sensitivity of {Large Language Model} Leaderboards},
  booktitle = {Proceedings of the 62nd Annual Meeting of the Association for Computational Linguistics ({ACL} 2024)},
  pages     = {13787--13805},
  year      = {2024},
  month     = aug,
  address   = {Bangkok, Thailand},
  publisher = {Association for Computational Linguistics},
  doi       = {10.18653/v1/2024.acl-long.744},
  url       = {https://aclanthology.org/2024.acl-long.744/}
}

@inproceedings{trustaibenchmarks2025aies,
  author    = {Eriksson, Maria and Purificato, Erasmo and Noroozian, Arman and Vinagre, Jo\~{a}o and Chaslot, Guillaume and G\'{o}mez, Emilia and Fern\'{a}ndez-Llorca, David},
  title     = {Can We Trust {AI} Benchmarks? {An} Interdisciplinary Review of Current Issues in {AI} Evaluation},
  booktitle = {Proceedings of the {AAAI}/{ACM} Conference on {AI}, Ethics, and Society},
  volume    = {8},
  number    = {1},
  pages     = {850--864},
  year      = {2025},
  doi       = {10.1609/aies.v8i1.36595},
  url       = {https://doi.org/10.1609/aies.v8i1.36595}
}

@article{SoTA_benchmark_culture2025springer,
  author    = {Campolo, Alexander},
  title     = {State-of-the-Art: The Temporal Order of Benchmarking Culture},
  journal   = {Digital Society},
  volume    = {4},
  number    = {2},
  pages     = {35},
  year      = {2025},
  publisher = {Springer},
  doi       = {10.1007/s44206-025-00190-x},
  url       = {https://doi.org/10.1007/s44206-025-00190-x}
}

@article{russakovsky2015imagenet,
  author  = {Russakovsky, Olga and Deng, Jia and Su, Hao and Krause, Jonathan and Satheesh, Sanjeev and Ma, Sean and Huang, Zhiheng and Karpathy, Andrej and Khosla, Aditya and Bernstein, Michael S. and Berg, Alexander C. and Fei-Fei, Li},
  title   = {{ImageNet} Large Scale Visual Recognition Challenge},
  journal = {International Journal of Computer Vision},
  volume  = {115},
  number  = {3},
  pages   = {211--252},
  year    = {2015},
  doi     = {10.1007/s11263-015-0816-y},
  url     = {https://doi.org/10.1007/s11263-015-0816-y}
}

@inproceedings{deng2009imagenet,
  author    = {Deng, Jia and Dong, Wei and Socher, Richard and Li, Li-Jia and Li, Kai and Fei-Fei, Li},
  title     = {{ImageNet}: A Large-Scale Hierarchical Image Database},
  booktitle = {2009 {IEEE} Conference on Computer Vision and Pattern Recognition},
  pages     = {248--255},
  year      = {2009},
  publisher = {{IEEE}},
  doi       = {10.1109/CVPR.2009.5206848},
  url       = {https://doi.org/10.1109/CVPR.2009.5206848}
}

@inproceedings{wang2018glue,
  author    = {Wang, Alex and Singh, Amanpreet and Michael, Julian and Hill, Felix and Levy, Omer and Bowman, Samuel R.},
  title     = {{GLUE}: A Multi-Task Benchmark and Analysis Platform for Natural Language Understanding},
  booktitle = {Proceedings of the 2018 {EMNLP} Workshop {BlackboxNLP}: Analyzing and Interpreting Neural Networks for {NLP}},
  pages     = {353--355},
  year      = {2018},
  address   = {Brussels, Belgium},
  publisher = {Association for Computational Linguistics},
  doi       = {10.18653/v1/W18-5446},
  url       = {https://aclanthology.org/W18-5446/}
}

@inproceedings{liao2021arewelearningyet,
  author    = {Liao, Thomas and Taori, Rohan and Raji, Inioluwa Deborah and Schmidt, Ludwig},
  title     = {Are We Learning Yet? {A} Meta Review of Evaluation Failures Across Machine Learning},
  booktitle = {Proceedings of the Neural Information Processing Systems Track on Datasets and Benchmarks},
  volume    = {1},
  year      = {2021},
  url       = {https://datasets-benchmarks-proceedings.neurips.cc/paper_files/paper/2021/file/757b505cfd34c64c85ca5b5690ee5293-Paper-round2.pdf}
}

@inproceedings{daynauth2025rankingunraveled,
  author    = {Daynauth, Roland and Clarke, Christopher and Flautner, Krisztian and Tang, Lingjia and Mars, Jason},
  title     = {Ranking Unraveled: Recipes for {LLM} Rankings in Head-to-Head {AI} Combat},
  booktitle = {Proceedings of the 63rd Annual Meeting of the Association for Computational Linguistics ({ACL} 2025)},
  pages     = {26078--26091},
  year      = {2025},
  address   = {Vienna, Austria},
  publisher = {Association for Computational Linguistics},
  doi       = {10.18653/v1/2025.acl-long.1265},
  url       = {https://aclanthology.org/2025.acl-long.1265/}
}

@inproceedings{chatbotarena2024icml,
  author    = {Chiang, Wei-Lin and Zheng, Lianmin and Sheng, Ying and Angelopoulos, Anastasios Nikolas and Li, Tianle and Li, Dacheng and Zhu, Banghua and Zhang, Hao and Jordan, Michael and Gonzalez, Joseph E. and Stoica, Ion},
  title     = {Chatbot Arena: An Open Platform for Evaluating {LLM}s by Human Preference},
  booktitle = {Proceedings of the 41st International Conference on Machine Learning},
  series    = {Proceedings of Machine Learning Research},
  volume    = {235},
  pages     = {8359--8388},
  year      = {2024},
  publisher = {{PMLR}},
  url       = {https://proceedings.mlr.press/v235/chiang24b.html}
}

@article{helm2023tmlr,
  author  = {Liang, Percy and Bommasani, Rishi and Lee, Tony and Tsipras, Dimitris and Soylu, Dilara and Yasunaga, Michihiro and Zhang, Yian and Narayanan, Deepak and Wu, Yuhuai and Kumar, Ananya and Newman, Benjamin and Yuan, Binhang and Yan, Bobby and Zhang, Ce and Cosgrove, Christian and Manning, Christopher D. and R\'{e}, Christopher and Acosta-Navas, Diana and Hudson, Drew A. and Zelikman, Eric and Durmus, Esin and Ladhak, Faisal and Rong, Frieda and Ren, Hongyu and Yao, Huaxiu and Wang, Jue and Santhanam, Keshav and Orr, Laurel and Zheng, Lucia and Yuksekgonul, Mert and Suzgun, Mirac and Kim, Nathan and Guha, Neel and Chatterji, Niladri and Khattab, Omar and Henderson, Peter and Huang, Qian and Chi, Ryan and Xie, Sang Michael and Santurkar, Shibani and Ganguli, Surya and Hashimoto, Tatsunori and Icard, Thomas and Zhang, Tianyi and Chaudhary, Vishrav and Wang, William and Li, Xuechen and Mai, Yifan and Zhang, Yuhui and Koreeda, Yuta},
  title   = {Holistic Evaluation of Language Models},
  journal = {Transactions on Machine Learning Research},
  issn    = {2835-8856},
  year    = {2023},
  url     = {https://openreview.net/forum?id=iO4LZibEqW}
}

@inproceedings{reducedreusedrecycled2021neurips,
  author    = {Koch, Bernard and Denton, Emily and Hanna, Alex and Foster, Jacob Gates},
  title     = {Reduced, Reused and Recycled: The Life of a Dataset in Machine Learning Research},
  booktitle = {Proceedings of the Neural Information Processing Systems Track on Datasets and Benchmarks},
  volume    = {1},
  year      = {2021},
  url       = {https://openreview.net/forum?id=zNQBIBKJRkd}
}

@inproceedings{measuringwhatmatters2025neurips,
    title     = {Measuring what Matters: Construct Validity in Large Language Model Benchmarks},
    author    = {Andrew M. Bean and Ryan Othniel Kearns and Angelika Romanou and Franziska Sofia Hafner and Harry Mayne and Jan Batzner and Negar Foroutan and Chris Schmitz and Karolina Korgul and Hunar Batra and Oishi Deb and Emma Beharry and Cornelius Emde and Thomas Foster and Anna Gausen and Mar{\'\i}a Grandury and Simeng Han and Valentin Hofmann and Lujain Ibrahim and Hazel Kim and Hannah Rose Kirk and Fangru Lin and Gabrielle Kaili-May Liu and Lennart Luettgau and Jabez Magomere and Jonathan Rystr{\o}m and Anna Sotnikova and Yushi Yang and Yilun Zhao and Adel Bibi and Antoine Bosselut and Ronald Clark and Arman Cohan and Jakob Nicolaus Foerster and Yarin Gal and Scott A. Hale and Inioluwa Deborah Raji and Christopher Summerfield and Philip Torr and Cozmin Ududec and Luc Rocher and Adam Mahdi},
    booktitle = {The Thirty-ninth Annual Conference on Neural Information Processing Systems Datasets and Benchmarks Track},
    year      = {2026},
    url       = {https://openreview.net/forum?id=mdA5lVvNcU}
}

@inproceedings{voterigging2025icml,
  author    = {Min, Rui and Pang, Tianyu and Du, Chao and Liu, Qian and Cheng, Minhao and Lin, Min},
  title     = {Improving Your Model Ranking on Chatbot Arena by Vote Rigging},
  booktitle = {Proceedings of the 42nd International Conference on Machine Learning},
  series    = {Proceedings of Machine Learning Research},
  volume    = {267},
  pages     = {44252--44271},
  year      = {2025},
  publisher = {{PMLR}},
  url       = {https://proceedings.mlr.press/v267/min25a.html}
}

@inproceedings{droppinghandful2026iclr,
  author    = {Huang, Jenny Y. and Shen, Yunyi and Wei, Dennis and Broderick, Tamara},
  title     = {Dropping Just a Handful of Preferences Can Change Top {Large Language Model} Rankings},
  booktitle = {The Fourteenth International Conference on Learning Representations},
  year      = {2026},
  url       = {https://openreview.net/forum?id=jNiEMDsRgc}
}

@inproceedings{aligningleaderboardwithhumanpreference2025naacl,
  author    = {Gao, Mingqi and Liu, Yixin and Hu, Xinyu and Wan, Xiaojun and Bragg, Jonathan and Cohan, Arman},
  title     = {Re-evaluating Automatic {LLM} System Ranking for Alignment with Human Preference},
  booktitle = {Findings of the Association for Computational Linguistics: {NAACL} 2025},
  pages     = {4605--4629},
  year      = {2025},
  month     = apr,
  address   = {Albuquerque, New Mexico},
  publisher = {Association for Computational Linguistics},
  doi       = {10.18653/v1/2025.findings-naacl.260},
  url       = {https://aclanthology.org/2025.findings-naacl.260/},
  isbn      = {979-8-89176-195-7}
}

@inproceedings{leaderboardillusion2025neurips,
  author    = {Singh, Shivalika and Nan, Yiyang and Wang, Alex and D'Souza, Daniel and Kapoor, Sayash and {\"U}stun, Ahmet and Koyejo, Sanmi and Deng, Yuntian and Longpre, Shayne and Smith, Noah A. and Ermis, Beyza and Fadaee, Marzieh and Hooker, Sara},
  title     = {The Leaderboard Illusion},
  booktitle = {The Thirty-ninth Annual Conference on Neural Information Processing Systems Datasets and Benchmarks Track},
  year      = {2025},
  url       = {https://openreview.net/forum?id=4Ae8edNqm0}
}

@inproceedings{amelo2025icml,
  author    = {Liu, Zirui and Li, Jiatong and Zhuang, Yan and Liu, Qi and Shen, Shuanghong and Ouyang, Jie and Cheng, Mingyue and Wang, Shijin},
  title     = {am-{ELO}: A Stable Framework for Arena-based {LLM} Evaluation},
  booktitle = {Proceedings of the 42nd International Conference on Machine Learning},
  series    = {Proceedings of Machine Learning Research},
  volume    = {267},
  pages     = {38857--38868},
  year      = {2025},
  publisher = {{PMLR}},
  url       = {https://proceedings.mlr.press/v267/liu25ak.html}
}

@article{diagnoseBTframework2022,
  author  = {Wu, Weichen and Niezink, Nynke and Junker, Brian},
  title   = {A Diagnostic Framework for the {Bradley--Terry} Model},
  journal = {Journal of the Royal Statistical Society Series A: Statistics in Society},
  volume  = {185},
  number  = {Supplement 2},
  pages   = {S461--S484},
  year    = {2022},
  month   = nov,
  issn    = {0964-1998},
  doi     = {10.1111/rssa.12959},
  url     = {https://doi.org/10.1111/rssa.12959}
}

@article{uncertaintyquantificationBT2023ima,
  author  = {Gao, Chao and Shen, Yandi and Zhang, Anderson Y.},
  title   = {Uncertainty quantification in the {Bradley--Terry--Luce} model},
  journal = {Information and Inference: A Journal of the {IMA}},
  volume  = {12},
  number  = {2},
  pages   = {1073--1140},
  year    = {2023},
  month   = jun,
  doi     = {10.1093/imaiai/iaac032},
  url     = {https://doi.org/10.1093/imaiai/iaac032}
}

@article{bigbench2023tmlr,
  author  = {Srivastava, Aarohi and others},
  title   = {Beyond the Imitation Game: Quantifying and extrapolating the capabilities of language models},
  journal = {Transactions on Machine Learning Research},
  issn    = {2835-8856},
  year    = {2023},
  url     = {https://openreview.net/forum?id=uyTL5Bvosj}
}

@article{Braun2006,
  author    = {Braun, Virginia and Clarke, Victoria},
  title     = {Using thematic analysis in psychology},
  journal   = {Qualitative Research in Psychology},
  volume    = {3},
  number    = {2},
  pages     = {77--101},
  year      = {2006},
  publisher = {Routledge},
  doi       = {10.1191/1478088706qp063oa},
  url       = {https://doi.org/10.1191/1478088706qp063oa}
}

@article{reflexivethematicanalysis2019qrseh,
  author    = {Braun, Virginia and Clarke, Victoria},
  title     = {Reflecting on reflexive thematic analysis},
  journal   = {Qualitative Research in Sport, Exercise and Health},
  volume    = {11},
  number    = {4},
  pages     = {589--597},
  year      = {2019},
  publisher = {Routledge},
  doi       = {10.1080/2159676X.2019.1628806},
  url       = {https://doi.org/10.1080/2159676X.2019.1628806}
}

@article{Braun2021,
  author    = {Braun, Virginia and Clarke, Victoria},
  title     = {One size fits all? What counts as quality practice in (reflexive) thematic analysis?},
  journal   = {Qualitative Research in Psychology},
  volume    = {18},
  number    = {3},
  pages     = {328--352},
  year      = {2021},
  publisher = {Routledge},
  doi       = {10.1080/14780887.2020.1769238},
  url       = {https://doi.org/10.1080/14780887.2020.1769238}
}

@misc{openllmleaderboard2023,
  author       = {Beeching, Edward and Fourrier, Cl\'{e}mentine and Habib, Nathan and Han, Sheon and Lambert, Nathan and Rajani, Nazneen and Sanseviero, Omar and Tunstall, Lewis and Wolf, Thomas},
  title        = {Open {LLM} Leaderboard},
  year         = {2023},
  publisher    = {Hugging Face},
  howpublished = {\url{https://huggingface.co/spaces/open-llm-leaderboard/open_llm_leaderboard}},
  note         = {Accessed: 2026-01-10. You can find the snapshot of this leaderboard at \url{https://web.archive.org/web/20260301031113/https://huggingface.co/spaces/open-llm-leaderboard/open_llm_leaderboard}}
}

@article{strathern1997,
  author  = {Strathern, Marilyn},
  title   = {`Improving ratings': audit in the {British} {University} system},
  journal = {European Review},
  volume  = {5},
  number  = {3},
  pages   = {305--321},
  year    = {1997},
  doi     = {10.1017/S1062798700002660},
  url     = {https://doi.org/10.1017/S1062798700002660}
}

@article{simon1955,
  author  = {Simon, Herbert A.},
  title   = {A Behavioral Model of Rational Choice},
  journal = {The Quarterly Journal of Economics},
  volume  = {69},
  number  = {1},
  pages   = {99--118},
  year    = {1955},
  doi     = {10.2307/1884852},
  url     = {https://doi.org/10.2307/1884852}
}

@inproceedings{hendrycks2021mmlu,
  author    = {Hendrycks, Dan and Burns, Collin and Basart, Steven and Zou, Andy and Mazeika, Mantas and Song, Dawn and Steinhardt, Jacob},
  title     = {Measuring Massive Multitask Language Understanding},
  booktitle = {International Conference on Learning Representations},
  year      = {2021},
  address   = {Online},
  publisher = {OpenReview.net},
  url       = {https://openreview.net/forum?id=d7KBjmI3GmQ}
}

@inproceedings{ethayarajh2020utility,
  author    = {Ethayarajh, Kawin and Jurafsky, Dan},
  title     = {Utility is in the Eye of the User: A Critique of {NLP} Leaderboards},
  booktitle = {Proceedings of the 2020 Conference on Empirical Methods in Natural Language Processing ({EMNLP})},
  pages     = {4846--4853},
  year      = {2020},
  address   = {Online},
  publisher = {Association for Computational Linguistics},
  doi       = {10.18653/v1/2020.emnlp-main.393},
  url       = {https://aclanthology.org/2020.emnlp-main.393/}
}

@inproceedings{strubell2019energy,
  author    = {Strubell, Emma and Ganesh, Ananya and McCallum, Andrew},
  title     = {Energy and Policy Considerations for Deep Learning in {NLP}},
  booktitle = {Proceedings of the 57th Annual Meeting of the Association for Computational Linguistics ({ACL} 2019)},
  pages     = {3645--3650},
  year      = {2019},
  address   = {Florence, Italy},
  publisher = {Association for Computational Linguistics},
  doi       = {10.18653/v1/P19-1355},
  url       = {https://aclanthology.org/P19-1355/}
}

@article{schwartz2020green,
  author  = {Schwartz, Roy and Dodge, Jesse and Smith, Noah A. and Etzioni, Oren},
  title   = {Green {AI}},
  journal = {Communications of the {ACM}},
  volume  = {63},
  number  = {12},
  pages   = {54--63},
  year    = {2020},
  month   = dec,
  doi     = {10.1145/3381831},
  url     = {https://doi.org/10.1145/3381831}
}

@article{lipton2019troubling,
  author  = {Lipton, Zachary C. and Steinhardt, Jacob},
  title   = {Troubling Trends in Machine Learning Scholarship},
  journal = {{ACM} Queue},
  volume  = {17},
  number  = {1},
  pages   = {45--77},
  year    = {2019},
  doi     = {10.1145/3317287.3328534},
  url     = {https://queue.acm.org/detail.cfm?id=3328534}
}

@book{rogers2003diffusion,
  author    = {Rogers, Everett M.},
  title     = {Diffusion of Innovations},
  edition   = {5},
  publisher = {Free Press},
  address   = {New York, NY},
  year      = {2003},
  isbn      = {9780743222099},
  url       = {https://www.simonandschuster.com/books/Diffusion-of-Innovations-5th-Edition/Everett-M-Rogers/9780743222099}
}

@inproceedings{mitchell2019modelcards,
  author    = {Mitchell, Margaret and Wu, Simone and Zaldivar, Andrew and Barnes, Parker and Vasserman, Lucy and Hutchinson, Ben and Spitzer, Elena and Raji, Inioluwa Deborah and Gebru, Timnit},
  title     = {Model Cards for Model Reporting},
  booktitle = {Proceedings of the Conference on Fairness, Accountability, and Transparency ({FAT*} '19)},
  pages     = {220--229},
  year      = {2019},
  address   = {Atlanta, GA, USA},
  publisher = {Association for Computing Machinery},
  doi       = {10.1145/3287560.3287596},
  url       = {https://doi.org/10.1145/3287560.3287596}
}

@article{guest2006interviews,
  author  = {Guest, Greg and Bunce, Arwen and Johnson, Laura},
  title   = {How Many Interviews Are Enough? An Experiment with Data Saturation and Variability},
  journal = {Field Methods},
  volume  = {18},
  number  = {1},
  pages   = {59--82},
  year    = {2006},
  doi     = {10.1177/1525822X05279903},
  url     = {https://doi.org/10.1177/1525822X05279903}
}

@inproceedings{eslami2015invisiblealgorithms,
  author    = {Eslami, Motahhare and Rickman, Aimee and Vaccaro, Kristen and Aleyasen, Amirhossein and Vuong, Andy and Karahalios, Karrie and Hamilton, Kevin and Sandvig, Christian},
  title     = {``{I} always assumed that {I} wasn't really that close to [her]'': Reasoning about Invisible Algorithms in News Feeds},
  booktitle = {Proceedings of the 33rd Annual {ACM} Conference on Human Factors in Computing Systems ({CHI} '15)},
  pages     = {153--162},
  year      = {2015},
  publisher = {{ACM}},
  doi       = {10.1145/2702123.2702556},
  url       = {https://doi.org/10.1145/2702123.2702556}
}

@inproceedings{rader2015userbeliefs,
  author    = {Rader, Emilee and Gray, Rebecca},
  title     = {Understanding User Beliefs About Algorithmic Curation in the {Facebook} News Feed},
  booktitle = {Proceedings of the 33rd Annual {ACM} Conference on Human Factors in Computing Systems ({CHI} '15)},
  pages     = {173--182},
  year      = {2015},
  publisher = {{ACM}},
  doi       = {10.1145/2702123.2702174},
  url       = {https://doi.org/10.1145/2702123.2702174}
}

\appendix
\section{Interview Guide}
\label{sec:appendix:interview_guide}

The interview was structured into eight phases. The numbered items below were treated as anchor questions and were asked of every participant; the follow-up probes (marked with `` '') were used adaptively depending on the participant's responses. The total interview length was 30--45 minutes; the relative weight of each phase varied across participants and is not reported here.

\paragraph{Phase 1: Introduction and consent.}
The interviewer paraphrased the information letter, restated the voluntariness of participation, confirmed verbal consent for audio recording, and only then started the recorder. Participants were reminded that there were no right or wrong answers and that they could skip any question.

\paragraph{Phase 2: Warm-up and research context.}
\begin{enumerate}
    \item \textit{``Tell me a bit about your current research. What are you working on and what excites you about it?''}
    \item \textit{``How do LLMs fit into your research day-to-day, e.g.\ as tools, as objects of study, or something else?''}
    \begin{itemize}
        \item   What models do you use most often? How did you end up with those?
        \item   Has the role of LLMs in your work changed over the past year or two?
    \end{itemize}
\end{enumerate}

\paragraph{Phase 3: Model selection and workflow (RQ1).}
\begin{enumerate}
    \setcounter{enumi}{2}
    \item \textit{``Tell me about the last time you were choosing a model for a project. What did that process actually look like?''}
    \begin{itemize}
        \item   Where did you start? What sources of information did you consult?
        \item   Did you look at any rankings, benchmarks, or leaderboards during that process?
        \item   How much of the decision was based on numbers vs.\ recommendations from colleagues or your own intuition?
    \end{itemize}
    \item \textit{``When you hear that a model is `state-of-the-art,' what does that mean to you? Where does that claim usually come from?''}
    \begin{itemize}
        \item   Is that a label you take at face value, or do you unpack it?
        \item   Does it matter who is making the claim (a company, a leaderboard, a peer)?
    \end{itemize}
    \item \textit{``Have you ever chosen a lower-ranked model over a higher-ranked one? What was the reason?''}
    \begin{itemize}
        \item   Was it cost, latency, API availability, task fit, or something else?
        \item   How did you justify that choice to yourself or to collaborators/reviewers?
    \end{itemize}
    \item \textit{``Do you check leaderboards once and move on, or do you monitor them over time? What happens if a model you're using drops in rankings mid-project?''}
    \begin{itemize}
        \item   Has that ever actually happened to you?
        \item   Would it change your research plans?
    \end{itemize}
\end{enumerate}

\paragraph{Phase 4: Leaderboard encounters and use (RQ1 + RQ2).}
\begin{enumerate}
    \setcounter{enumi}{6}
    \item \textit{``Can you walk me through a specific time when a leaderboard result or benchmark score directly influenced a decision you made in your research?''}
    \begin{itemize}
        \item   What was the decision? Which leaderboard or benchmark was it?
        \item   Looking back, did the leaderboard steer you in the right direction?
    \end{itemize}
    \item \textit{``Do you see meaningful differences between arena-based rankings (e.g., Chatbot Arena, where humans vote on outputs) and fixed benchmark leaderboards like MMLU or HELM?''}
    \begin{itemize}
        \item   Do you trust one type more than the other? Why?
        \item   Do you use them for different purposes?
        \item   Are you aware of concerns about vote manipulation in arena-based platforms?
    \end{itemize}
    \item \textit{``Have you ever cited a leaderboard ranking in a paper or presentation to justify a model choice?''}
    \begin{itemize}
        \item   How did that feel? Was it a strong justification or more of a shorthand?
        \item   Do you see others doing this in your field?
    \end{itemize}
\end{enumerate}

\paragraph{Phase 5: Trust, skepticism, and limitations (RQ2).}
\begin{enumerate}
    \setcounter{enumi}{9}
    \item \textit{``Tell me about a time a leaderboard result surprised you, either because it contradicted your experience or because it didn't match what you expected.''}
    \begin{itemize}
        \item   What did you do with that surprise? Did it change your view of the leaderboard, or of the model?
        \item   Did you investigate further, or just move on?
    \end{itemize}
    \item \textit{``In your view, is there such a thing as the `best' LLM? Or is that always task- and context-dependent?''}
    \begin{itemize}
        \item   If it's context-dependent, do leaderboards do a good enough job of representing that?
    \end{itemize}
    \item \textit{``What do leaderboards fail to capture that actually matters for your work?''}
    \begin{itemize}
        \item   Are there dimensions (e.g., cost, fairness, reliability, or user experience) that you wish were represented?
        \item   Is the missing piece something quantifiable, or something harder to measure?
    \end{itemize}
    \item \textit{``Are you aware of technical concerns about leaderboard fragility, for instance the fact that small changes to benchmark data or ranking algorithms can flip model orderings?''}
    \begin{itemize}
        \item   If yes: does that knowledge change how you use them?
        \item   If no: [the interviewer briefly explained recent findings] does hearing that change your perspective?
    \end{itemize}
\end{enumerate}

\paragraph{Phase 6: Social and disciplinary norms (RQ1 + RQ2).}
\begin{enumerate}
    \setcounter{enumi}{13}
    \item \textit{``Have you ever felt pressure (from reviewers, collaborators, or just the culture of your field) to use or compare against a top-ranked model? What was that experience like?''}
    \begin{itemize}
        \item   Was the pressure explicit (a reviewer comment) or implicit (just understood)?
        \item   Did you push back, or comply? What happened?
    \end{itemize}
    \item \textit{``How do reviewers in your field typically react to model choices? Has a reviewer ever questioned why you didn't use a top-ranked model?''}
    \begin{itemize}
        \item   Do you think this is a healthy norm, or does it distort research?
    \end{itemize}
\end{enumerate}

\paragraph{Phase 7: Imagining alternatives.}
\begin{enumerate}
    \setcounter{enumi}{15}
    \item \textit{``If you could redesign the way LLMs are evaluated and compared, what would you change?''}
    \begin{itemize}
        \item   Would you change what's measured, how it's presented, who controls it, or something else?
        \item   Are there evaluation practices from your own subfield that you think leaderboards could learn from?
    \end{itemize}
    \item \textit{``What information do you wish leaderboards provided but currently don't?''}
    \begin{itemize}
        \item   For example: confidence intervals, cost-per-query, task-specific breakdowns, fairness metrics, reproducibility information?
        \item   Would that change how you use them?
    \end{itemize}
\end{enumerate}

\paragraph{Phase 8: Wrap-up.}
\begin{enumerate}
    \setcounter{enumi}{17}
    \item \textit{``Is there anything about your experience with leaderboards or benchmarks that we haven't touched on that you think is important?''}
    \item \textit{``If you had to describe your overall relationship with LLM leaderboards in one sentence, what would you say?''}
\end{enumerate}

\paragraph{Interviewer practices.}
The interviewer followed established practices for semi-structured interviewing in HCI qualitative research: critical-incident probing (asking participants to walk through a specific past event rather than offer generalized opinions), leaving 3--5 seconds of silence after a participant finished speaking to invite reflective additions, mirroring rather than leading (reflecting back the participant's framing rather than offering an interpretation), and adapting the order of probes to the energy of each conversation. Each participant's pre-interview questionnaire (Appendix~\ref{sec:appendix:questionnaire}) was reviewed immediately before the interview to tailor follow-ups to their subfield and self-reported leaderboard familiarity.

\paragraph{Interviewer notes.}
The primary researcher kept light field notes during each interview, and added any observations that could not be jotted down in the moment immediately after the recorder was stopped. The notes captured the participant's overall demeanour, particularly striking moments or phrases, anything that seemed specific to their subfield, and quick impressions of whether questions were landing as intended. These notes were informal interviewer aids; they did not serve the role of the analytic memos described in Section~\ref{sec:methods:analysis}, which were written separately during the coding phase.

\section{Pre-Interview Questionnaire}
\label{sec:appendix:questionnaire}

Prospective participants completed the following Google Form before the interview. Estimated completion time: \textasciitilde5 minutes. Required items are marked with an asterisk (\textsuperscript{*}). Items used to verify consent are reproduced verbatim from the form; multi-option items are listed as bullet points.

\paragraph{Consent.}
\begin{itemize}
    \item \textit{I confirm I have read the Participant Information Letter and Consent Form and understand what my participation involves.}\textsuperscript{*} (Yes / No)
    \item \textit{I understand that my participation is voluntary and that I may withdraw at any time without consequence.}\textsuperscript{*} (Yes / No)
    \item \textit{Do you agree to the use of anonymous quotations in any presentation, report or preprint that comes of this study?}\textsuperscript{*} (Yes / No)
    \item \textit{I consent to the audio recording of the interview session.}\textsuperscript{*} (Yes / No)
    \item \textit{Do you have any questions or concerns about participation before the interview?} (free text)
    \item \textit{I consent to participate in this research study, including the pre-interview questionnaire and the subsequent interview.}\textsuperscript{*} (Yes / No)
    \item \textit{Typed signature (full name).}\textsuperscript{*}
\end{itemize}

\paragraph{Section A: Demographics.}
\begin{itemize}
    \item Full name.\textsuperscript{*} (Stored separately from interview data and replaced with a participant ID during analysis.)
    \item Preferred pronouns. (optional)
    \item Email.\textsuperscript{*} (Used for scheduling and follow-up.)
    \item Current role.\textsuperscript{*} One of: Master's student / PhD student / Postdoctoral researcher / Faculty (Assistant / Associate / Full Professor) / Research scientist (industry or lab).
    \item Year in program or years since PhD. (optional, e.g., ``2nd year PhD'' or ``3 years post-PhD'')
    \item Department / Institution.\textsuperscript{*}
\end{itemize}

\paragraph{Section B: Research Background.}
\begin{itemize}
    \item Primary CS subfield.\textsuperscript{*} One of: NLP (including IR, CL, Speech analysis) / Systems (including OS, distributed systems, computer architecture, networking, databases, compilers, ML systems, and security and privacy) / HCI (including CSCW, human--AI interaction, and visualization and visual analytics) / Other.
    \item Brief description of your current research focus.\textsuperscript{*} (1--3 sentences or a few keywords.)
\end{itemize}

\paragraph{Section C: LLM and Leaderboard Exposure.}
\begin{itemize}
    \item How often do you interact with LLMs or chatbots in your research?\textsuperscript{*} One of: Daily / Weekly / Monthly / Rarely (a few times per year) / Never.
    \item Have you heard of any of the following leaderboards?\textsuperscript{*} (Select all that apply.) Chatbot Arena / Open LLM Leaderboard / HELM / BigBench / MMLU / SuperGLUE / None / Not Applicable / Other.
    \item Have you ever used a leaderboard or benchmark ranking to help you make a research decision (e.g., choosing a model to use or compare against, framing a research claim, or selecting a baseline)? One of: Yes / No / Unsure.
\end{itemize}

\paragraph{Section D: Logistics.}
\begin{itemize}
    \item Preferred interview format.\textsuperscript{*} (Select all that apply.) Video call / In-person (David R. Cheriton School of Computer Science, University of Waterloo).
    \item Availability windows.\textsuperscript{*} (Confirmation that the participant has scheduled a time slot.)
    \item Is there anything you'd like us to know before the interview? (e.g., topics to avoid, accessibility needs)
\end{itemize}

\end{document}